\documentclass[conference]{IEEEtran}
\IEEEoverridecommandlockouts

\usepackage{cite}
\usepackage{amsmath,amssymb,amsfonts}
\usepackage{algorithmic}
\usepackage{graphicx}
\usepackage{textcomp}
\usepackage{xcolor}
\usepackage{placeins}
\usepackage{comment}
\usepackage[colorlinks = true, citecolor = blue, urlcolor = gray, linkcolor = cyan]{hyperref}

\begin{document}

\title{Analyzing Narrative Processing in Large Language Models (LLMs): Using GPT4 to test BERT \\ 

\thanks{DFG (German Research Foundation, Deutsche Forschungsgemeinschaft)}
}

\author{\IEEEauthorblockN{\textbf{Patrick Krauss$^{1,*}$, Jannik Hösch$^{1,*}$, Claus Metzner$^{2}$, Andreas Maier$^{3}$, Peter Uhrig$^{4}$, Achim Schilling$^{1,2, \#}$}}
\IEEEauthorblockA{
$^{1}$ CCN Group, Pattern Recognition Lab, University Erlangen-Nürnberg, Germany\\
$^{2}$ Neuroscience Lab, University Hospital Erlangen, Germany\\
$^{3}$ Pattern Recognition Lab, University Erlangen-Nürnberg, Germany\\
$^{4}$ Digital Linguistics and Big Data, Department of English and American Studies, University Erlangen-Nürnberg, Germany\\
$^*$ authors contributed equally\\
$^\#$achim.schilling@fau.de}
}

\maketitle

\begin{abstract}
The ability to transmit and receive complex information via language is unique to humans and is the basis of traditions, culture and versatile social interactions. Through the disruptive introduction of transformer based large language models (LLMs) humans are not the only entity to “understand” and produce language any more. In the present study, we have performed the first steps to use LLMs as a model to understand fundamental mechanisms of language processing in neural networks, in order to make predictions and generate hypotheses on how the human brain does language processing. Thus, we have used ChatGPT to generate seven different stylistic variations of ten different narratives (Aesop’s fables). We used these stories as input for the open source LLM BERT and have analyzed the activation patterns of the hidden units of BERT using multi-dimensional scaling and cluster analysis. We found that the activation vectors of the hidden units cluster according to stylistic variations in earlier layers of BERT (1) than narrative content (4-5). Despite the fact that BERT consists of 12 identical building blocks that are stacked and trained on large text corpora, the different layers perform different tasks. This is a very useful model of the human brain, where self-similar structures, i.e. different areas of the cerebral cortex, can have different functions and are therefore well suited to processing language in a very efficient way. The proposed approach has the potential to open the black box of LLMs on the one hand, and might be a further step to unravel the neural processes underlying human language processing and cognition in general.
\end{abstract}


\section*{Introduction}
What are the minimum requirements of a system that is able to understand and produce language, and what are the underlying mechanisms? This question is not just scientifically interesting but also crucial in order to build and understand artificial general intelligence (AGI) systems \cite{ge2024openagi,bubeck2023sparks}. Thus, the focus on language processing makes sense as language is the basis of any kind of traditions, culture, or science \cite{carruthers2002cognitive, baumeister2002collective}. Although some animal species such as toothed whales have sophisticated  communication systems \cite{bergler2019orca, janik2014cetacean}, a complex language, which could be used to express any kind of idea is unique to the human species \cite{berwick2013evolution}. Thus, language is the ideal starting point when trying to understand the basis of general intelligence. Already in 1950, Alan Turing wondered in his seminal paper "Can Machines think?" \cite{turing1950computing} and invented an intelligence test for computers based on language skills which he originally called "Imitation Game", and which is nowadays known as the "Turing-test" \cite{turing1950computing, moor1976analysis, piccinini2000turing}. A simplified version of the Turing test proposes that a computer is intelligent, if it is able to pretend to be a human and fooling a real human examiner \cite{piccinini2000turing, sterrett2020genius}. Nevertheless, John Searle pointed out in 1980 that language processing is at most the basis of general intelligence but that fluent communication is not the same as real understanding. This insight he gained from a thought experiment called the "Chinese room", where he argued that fluent communication is possible by just looking up input messages and finding a matching output message. This procedure does not require any real understanding of the messages \cite{searle1980minds}. Nevertheless, language is the basis of any complex culture and intelligence in humans \cite{gentner2003language,pinker2003language}. Thus, a lot of effort is put in researching language processing. 

The topic is approached from two sides. On the one hand, cognitive neuroscience takes advantage of the improving neuroimaging techniques, to investigate language processing in the human brain \cite{de2017research}. The improved measurement devices in combination with the computing power and storage space of modern computer enable us to apply new measurement paradigms. Thus, artificial language stimuli based on the repetitive presentation of single words or sentences \cite{de2017research} are replaced by the presentation of natural language resp. speech (see e.g. \cite{schilling2021analysis, schuller2023attentional, schuller2024early, koelbl2023adaptive, garibyan2022neural}). Unfortunately, this data driven approach reaches its limits in explaining how language is processed in the human brain \cite{schilling2023predictive}. This fact is emphasized by the study of Jonas and Kording, who applied modern evaluation techniques used in cognitive neuroscience research with the purpose to explain the function of a simple microprocessor \cite{jonas2017could}. These experiments served as a cross-check of current neuroscience research. However, the authors showed that the methods and the approach were not sufficient to unravel the processing principles of the microprocessor. The conclusion that our current experimental paradigms and evaluation techniques are not well suited to understand a simple microprocessor, and thus are trivially also not well suited to investigate our brain, with its approx. 86 billion neurons \cite{herculano2009human, gerum2020sparsity}, suggests that we need a different approach to unravel the mechanisms of language processing in the brain.

Using artificial neural networks such as LLMs as model systems is very promising, because LLMs are already capable of performing complex cognitive tasks, and have the decisive advantage that their internal states and parameters can be read out with arbitrary precision at any time which is impossible for living brains. In a second step, these neural networks can be made more and more biologically plausible (see e.g. \cite{gerum2021integration, gerum2020sparsity, gerum2023leaky, stoll2023coincidence}), i.e. brain constrained \cite{pulvermuller2021biological}.
This approach is called Cognitive Computational Neuroscience (CCN, \cite{kriegeskorte2018cognitive}) and is applied in different areas of cognitive science (see e.g. \cite{stoewer2022neural, stoewer2023neural, stoewer2023multi,surendra2023word,schilling2023predictive}). Using LLMs as model systems to understand the principles of language processing has not just the advantage that one might learn something about the human brain but the research strand might also help to partly open the black box of machine learning and artificial intelligence \cite{castelvecchi2016can}.

In the present study, we have used the LLM BERT (Bidirectional Encoder Representations from Transformers)  \cite{devlin2018bert} as model system and used a second LLM (ChatGPT, \cite{brown2020language}) to generate suitable input data. BERT is a large language model developed and made open source by Google AI in 2018 and is based on the transformer architecture developed by Vaswani et al. one year earlier \cite{vaswani2017attention}. However, BERT consists only of the encoder part of the transformer network proposed by Vaswani and coworkers \cite{vaswani2017attention, devlin2018bert}.
In contrast to language models such as ELMO \cite{peters2018deep}, BERT is not a feature based approach but a fine-tuning approach, which means that BERT is trained in two steps: the pre-training step and the fine-tuning step. In the pre-training phase BERT is trained on huge unlabeled text corpora on the masked language task - where BERT has to predict randomly masked tokens- and the next sentence prediction tasks -where BERT must predict the following sentence \cite{devlin2018bert}. In the fine tuning step an additional layer is trained supervisedly on task specific labeled data \cite{devlin2018bert}. As input BERT gets a 512-dimensional vector (number of tokens) of 768-dimensional "WordPiece" embedding vectors \cite{wu2016google} representing the input words \cite{devlin2018bert}. The first token fed to the network is always a so called CLS (class) token, which can be regarded as storage unit of the context information \cite{devlin2018bert}. 
The BERT model itself consists of 12 stacked transformer (encoder) blocks (multi-attention head + dense layer) and each multi-attention head consists of 12 attention heads (for further information see \cite{devlin2018bert}). The whole model sums up to 110 Million trainable parameters \cite{devlin2018bert}.
In the present study, we investigated the hidden activations of the BERT units in order to draw conclusions on how and where BERT processes different writing styles as well as different narratives (semantic content). Thus, we extracted the hidden activations of all 12 transformer blocks and projected them using multi-dimensional scaling (MDS see also \cite{schilling2021quantifying, krauss2021analysis, krauss2018statistical}). Furthermore, we quantified the degree of separation (clustering) of the neuron activations using the label-free EDD-value (entropy of distance distribution, \cite{metzner2023beyond}) and the generalized discrimination value (GDV, \cite{schilling2021quantifying, metzner2023beyond, metzner2022classification, metzner2021sleep, schilling2022deep, traxdorf2019microstructure}). Surprisingly, we found that in the first transformer block, the neural activations cluster according to writing styles. In the later blocks (4-5) the clustering reflects narrative content. This finding provides evidence that there is some specialization in the neural network, which is not an intrinsic property of the architecture. We propose these findings as a starting point to further investigate language processing in LLMs in order to compare these findings to neural data on the one hand (as proposed by \cite{kriegeskorte2018cognitive}) and, on the other hand to open the black box of LLMs in order to generate explainable AI \cite{castelvecchi2016can}.  

\section*{Methods}
\subsection*{Narrative generation using ChatGPT}
As the number of input tokens of BERT is limited to 512, we have used short fables (Aesop's Fables: For the Instruction and Improvement of Youth, \cite{1834aesop}) as input tokens for BERT with a maximum of approx. 400 words. The original fables we used are shown in the first column of Table \ref{tab:inputdata}. 
To generate stylistic variations of these fables we used the ChatGPT chatbot interface with the GPT4 backend \cite{radford2019language}. The procedure of generating 6 different stylistic variations (7 including the original story) of the fables was achieved in a three step process (see Table \ref{tab:inputdata} second column):
\begin{enumerate}
    \item{Initial Prompt: First, the initial prompt containing the information on the general task is send to GPT4: "Hello I have a fable that I'd like to present in different narrative styles. My request is for you to creatively rephrase the fable into each of these styles. Please maintain the core message of the fable in each variation but feel free to be creative with the settings and styles."}
    \item{After that, the fable is presented to GPT4.}
    \item{The third prompt contains the information on the writing style GPT4 should use to rephrase the fable: "Rephrase the fable..." (Table \ref{tab:inputdata} third column). The rephrased fable is double-checked to be sure that the fable does fulfill all needed requirements (length, content etc.).}
\end{enumerate}

\begin{table*}[h!]
\begin{center}
\caption{Different narratives and variations in writing style generated using ChatGPT}
\begin{tabular}{cc||ccc}
Narrative (Fable) & Number & Style & Number & Prompt\\
\hline
The Town Mouse and the Country Mouse & 1 & Adventure Tale & 1 & Rephrase the fable into an adventure tale \\
The Owl and the Grashopper & 2 & Children Story & 2 & Rephrase the fable into a children's story\\
Mercury and the Woodman & 3 & Comedy Version & 3 & Rephrase the fable into a comedic version\\
The cat, the Cock and the Young Mouse & 4 & Historical Context & 4 & Rephrase the fable in a historical context\\
The Ass and the Lap Dog & 5 &  Mystery Story & 5 & Rephrase the fable into a mystery story\\
The Wolf and the House Dog & 6 & Original & 6 & -\\
The Fox without a Tail & 7 & Science-Fiction Setting & 7 & Rephrase the fable into a science-fiction setting\\
The Bees and Wasps and the Hornet & 8 & - & - &-\\
The Lark and Her Young Ones & 9 & - & - & -\\
The Cat and the Old Rat & 10 & - & - & -\\
\end{tabular}
\label{tab:inputdata}
\end{center}
\end{table*}

\subsection{Computational resources and programming libraries}
All simulations were run on a standard personal computer. The evaluation software was based on Python 3.10 \cite{oliphant2007python}. For matrix operations the numpy-library \cite{van2011numpy} was used and data visualization was done using MatplotLib \cite{hunter2007matplotlib} and the pylustrator library \cite{gerum2019pylustrator}. The dimensionality reduction through multi-dimensional scaling (MDS) and the principal component analysis (PCA) were done using the scikit learn library \cite{scikit-learn, sklearn_api}. 

\subsection*{Analysis of hidden activations and data visualization}
The created variations are presented to the non-fine-tuned base version of BERT (12 transformer blocks, 12 attention heads each, 110 Million parameters, \cite{devlin2018bert}), which has already been evaluated in earlier studies (see \cite{rahwan2019machine}). Thus, the fables were transformed to input token sequences using the BERT tokenizer \cite{wu2016google, devlin2018bert} and fed including the additional CLS token to BERT. In this study, we evaluated exclusively the CLS token across the different LLM attention layers. To access the neuron activations (values of CLS token) across the 12 transformer blocks we used the transformers library \cite{wolf2020transformers}. The 768-dimensional vectors (CLS token in each transformer block) for each story variation were then projected onto a 2D plane using multi-dimensional scaling (MDS). Additionally, the generalized discrimination value (GDV) value as well as the entropy of distances distribution (EDD) were calculated from the distance matrices derived from the high dimensional data in order to quantify the degree of clustering. For the exact formulas of EDD and GDV see \cite{schilling2021quantifying,metzner2023beyond}. The EDD is a value that quantifies whether points are distributed isotropically (value of 1) over a state-space, or whether they form clusters independently of any labels \cite{metzner2023beyond} (EDD smaller than 1).
The GDV, however, quantifies how well high-dimensional representations cluster according to given labels. Thus, this normalized value is independent of the dimensionality of the data and is 0 for no clustering and becomes more negative for clustering according to pre-defined labels.  
In this, study we used the semantic content of the stories (narratives) as well as the writing style as labels and colored the 2D representations of the CLS tokens according to these labels (Fig. \ref{fig:scatter-semantic} and Fig. \ref{fig:scatter-style}).

\section*{Results}
\begin{figure}[h!]
    \centering
    \includegraphics[width = 0.5\textwidth]{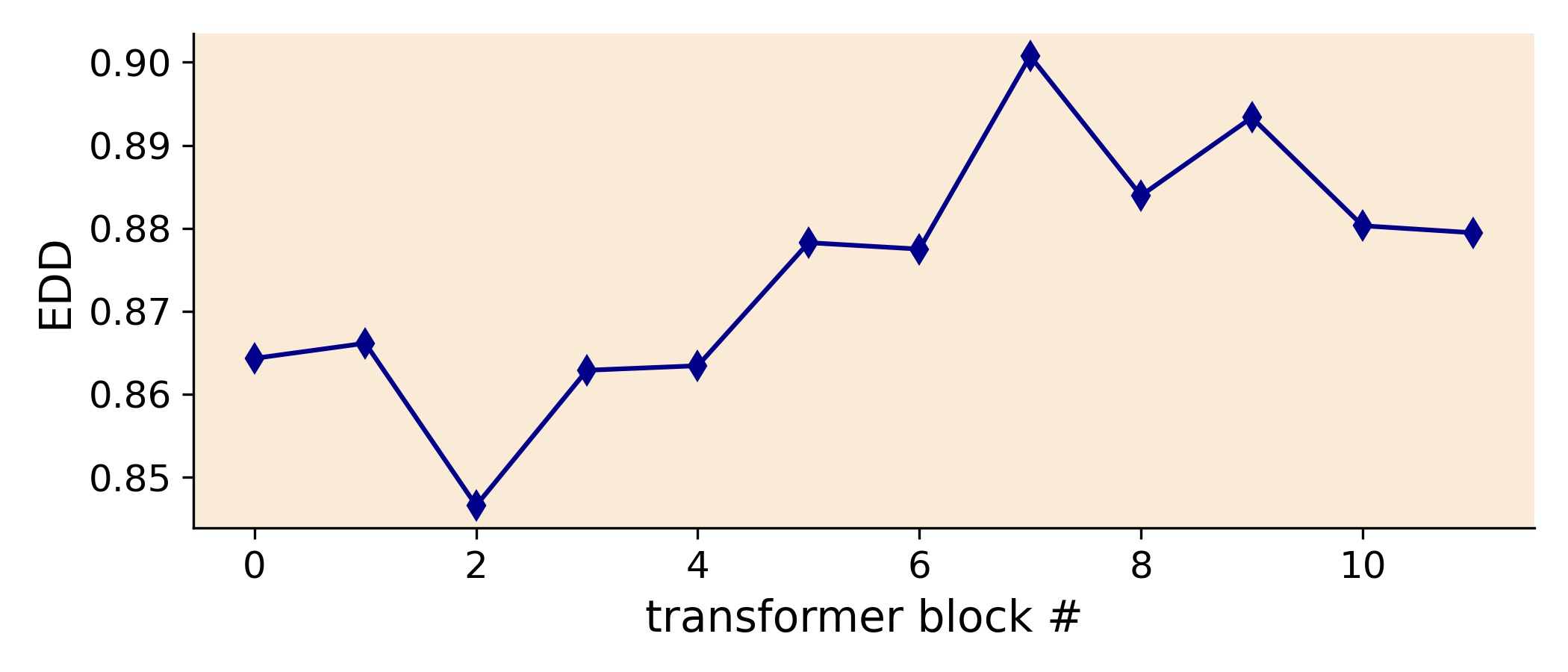}
    \caption{\textbf{Label-free measure of cluster formation}\newline The EDD  value shows that the cluster formation of the CLS token across the layers is minimal (no clear cluster evolve). The used CLS tokens (2D projections) are shown in Fig. \ref{fig:scatter-semantic} and Fig. \ref{fig:scatter-style}. Note that the labels (colors of the markers) play no role for the EDD value (see \cite{metzner2023beyond}).}
    \label{fig:EDD}
\end{figure}

\begin{figure*}[h!]
    \centering
    \includegraphics[width = \textwidth]{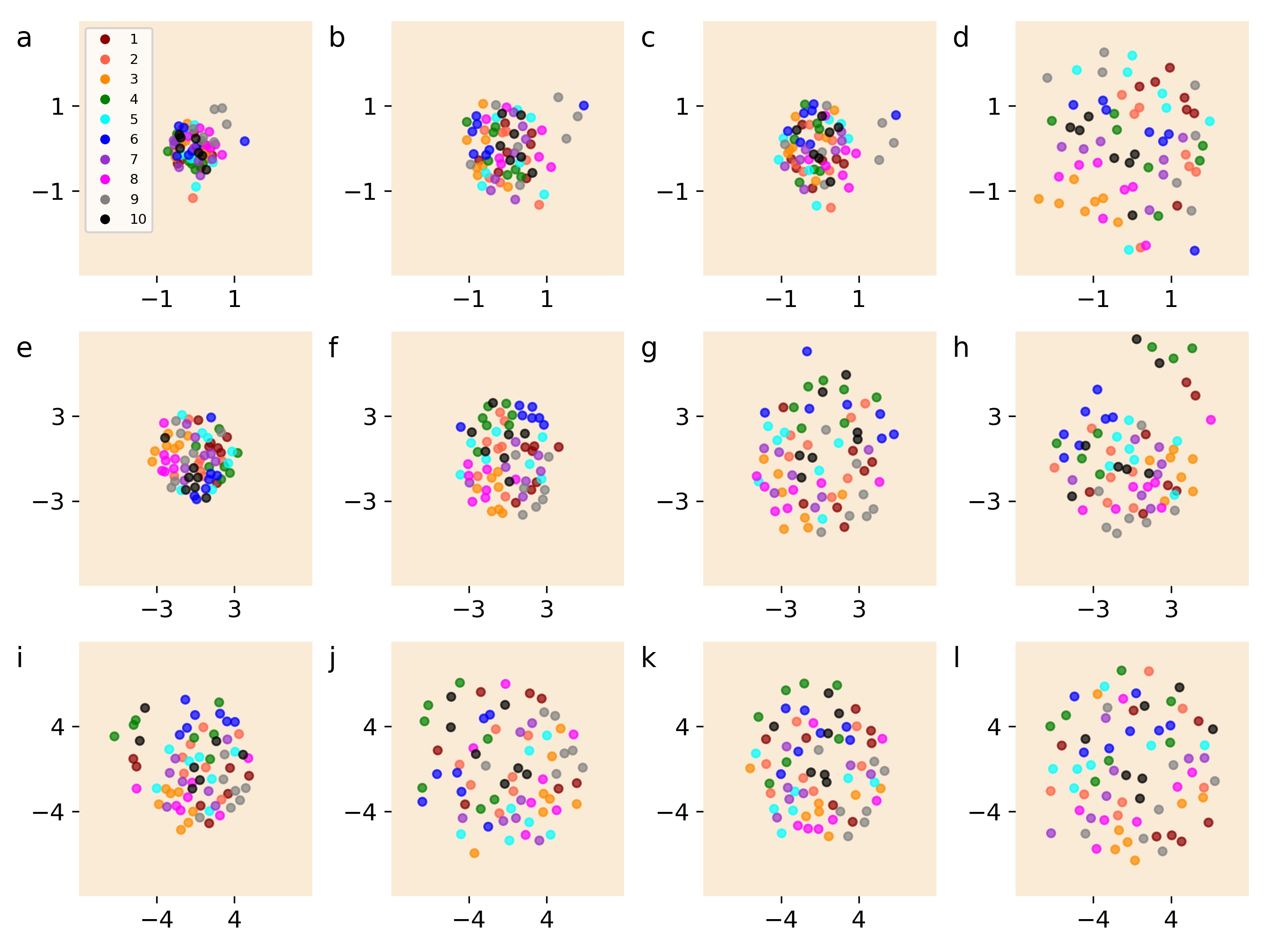}
    \caption{\textbf{2D projections of CLS token through all transformer blocks colored according to the narrative}\newline The plot shows the 2D projections of the CLS token of the 12 transformer blocks (block 1-12) projected with the MDS method. The colors correspond to the different narratives (fables). The numbers in the legend correspond to the numbers in Table \ref{tab:inputdata} 2nd column.}
    \label{fig:scatter-semantic}
\end{figure*}

\begin{figure*}[h!]
    \centering
    \includegraphics[width = \textwidth]{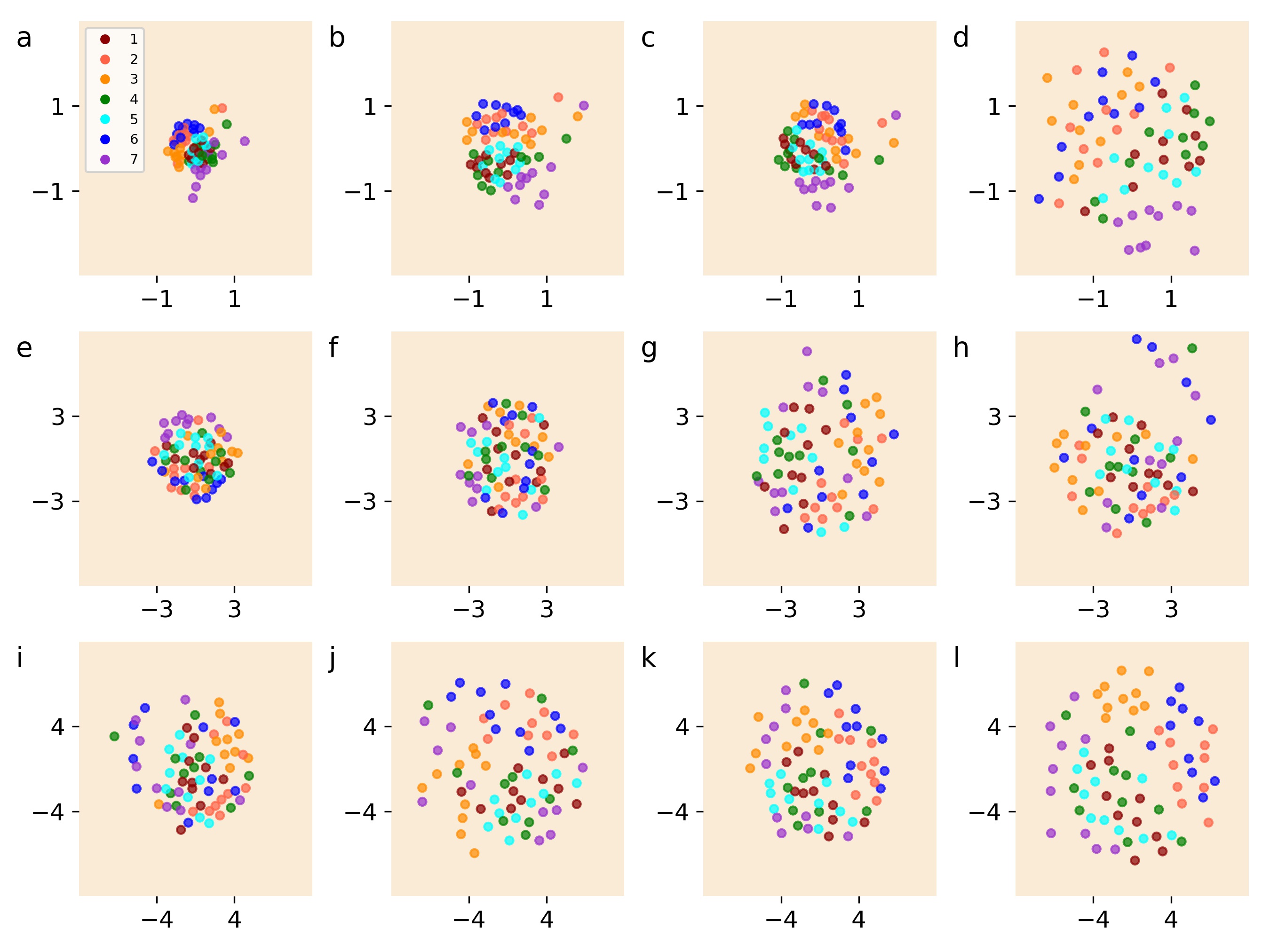}
    \caption{\textbf{2D projections of CLS token through all transformer blocks colored according to writing style}\newline The plot shows the MDS-projected CLS tokens of the 12 transformer blocks analogously to Fig \ref{fig:scatter-semantic}. The colors of the markers represent the different writing styles (see Table \ref{tab:inputdata} column 3 and 4 to find the according writing style of the numbers in the legend of the plot).}
    \label{fig:scatter-style}
\end{figure*}

The variance of the CLS tokens for different narratives and writing styles increases as a function of the transformer blocks (i.e. higher/later transformer blocks correspond to higher variance of CLS tokens), but no clear cluster of CLS tokens emerge. This is emphasized by the fact that the label-free EDD value \cite{metzner2023beyond} is near 1 ($>$ 0.847) for all transformer blocks (block 1-12, see Fig. \ref{fig:EDD}, input data used for EDD function is shown in Fig. \ref{fig:scatter-semantic} and \ref{fig:scatter-style}, EDD value is independent from pre-defined labels). Nevertheless, the fact that the EDD value is slightly smaller for early transformer blocks (1-4) demonstrates that the CLS tokens are forming weak, partial clusters in early transformer blocks. However, for later transformer blocks the EDD value and the average Euclidean distance between CLS tokens increase, which means that the CLS tokens are distributed more uniformly over the space of all possible CLS tokens. That does not mean that the CLS tokens are not ordered according to certain pre-defined labels such as content or writing style of the fables.  

\begin{figure}[h!]
    \centering
    \includegraphics[width = 0.5 \textwidth]{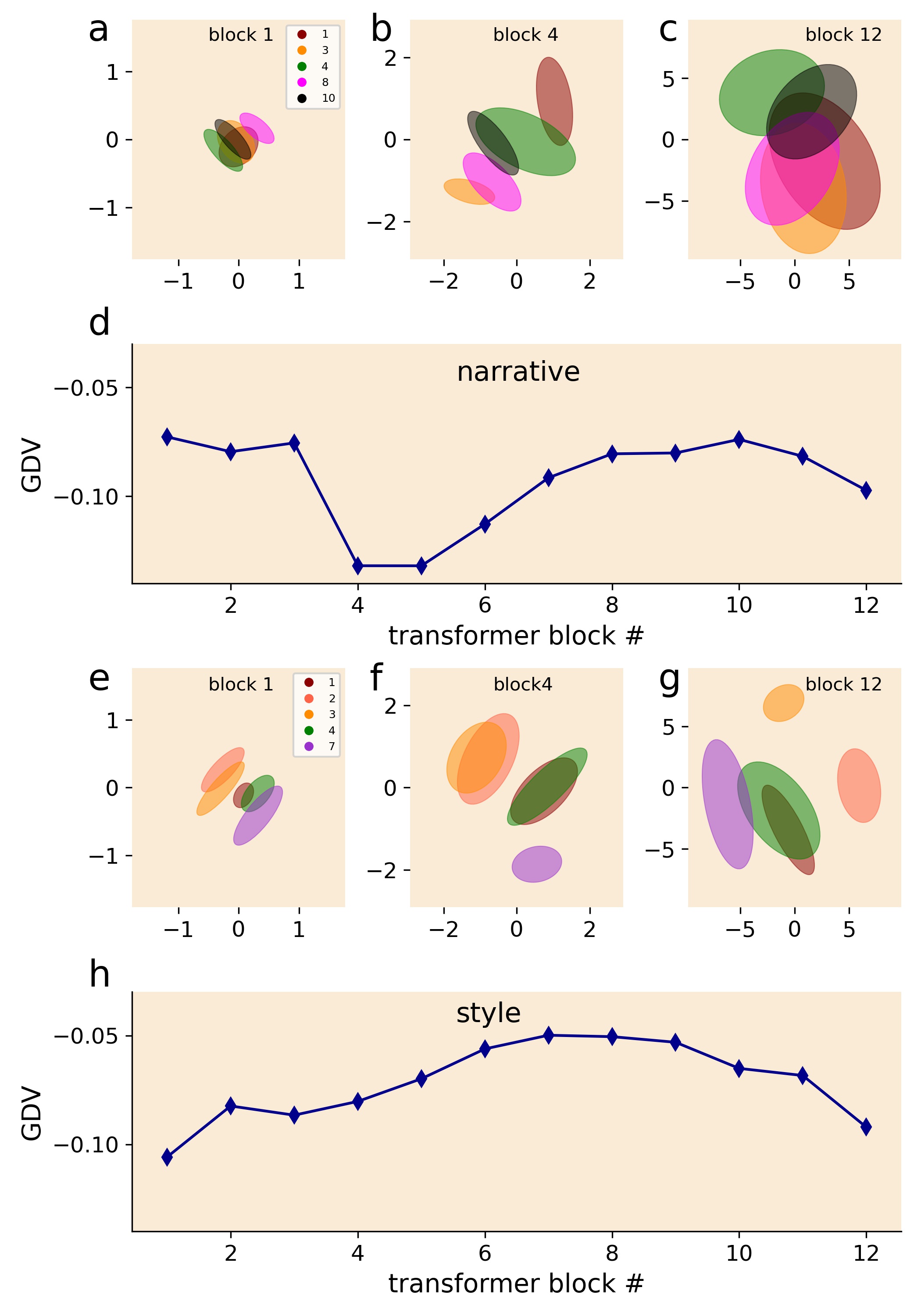}
    \caption{\textbf{GDV as a function of the layer resp. transformer block}\newline a-c: Shows the center of mass and the standard deviation of the CLS token along the principal axis plotted as colored ellipses for exemplary narratives and layers 1, 4, and 12. d: GDV as a function of the transformer block (layer). The CLS token vectors separate best (minimum of the GDV) according to the content of the story in block 4, 5. e-g: Same as a-c but for 5 exemplary writing styles; h: The GDV as a function of the writing style. The CLS vectors separate best (minimum of GDV) according to the writing style in block 1. The fact that the minimum of the GDV lies in different layers for different styles and different narratives illustrates that different layers of the transformer process language differently and are specialized on certain properties of language.}
    \label{fig:GDV}
\end{figure}

Indeed, the CLS tokens cluster according to the semantic content of the stories (10 different narratives in Table \ref{tab:inputdata} column 1). Thus, the  separation of the CLS tokens according to the semantic content (see Fig. \ref{fig:scatter-semantic}) is best after transformer blocks 4 and 5, indicated by a minimum of the layer-dependent GDV values (Fig. \ref{fig:GDV} d). Therefore, one could assume a central role of these transformer blocks in processing the plot of the input text. Indeed, the GDV decreases again in blocks 10-12, which might be caused by the fact that the last blocks merge the information on style and content.

In contrast to the findings on semantics shown above, the clustering of the CLS token according to the 7 different writing styles (see Table \ref{tab:inputdata}, for CLS projections see fig \ref{fig:scatter-style}) is significantly different. The minimum GDV value, which refers to the best clustering, is measured directly after the first transformer block (see Fig. \ref{fig:GDV} h). Starting from block 1 the GDV increases until block 7 indicating maximum class separation or clustering, and then decreases again (see Fig. \ref{fig:GDV} d).

The fact that semantic content and the writing style are represented in different transformer blocks suggest that certain transformer blocks are specialized on processing certain properties of language. However, in the last attention blocks of BERT, the CLS tokens cluster according to content as well as style, so that in both cases the GDV decreases again in the last transformer blocks (see Fig. \ref{fig:GDV} d, h).

\section*{Discussion}
\subsection*{Summary}
In the present study, we have shown the specialization of different transformer blocks of BERT \cite{devlin2018bert} on certain properties of natural language. We have used ChatGPT \cite{radford2019language} to generate 7 different stylistic variations of 10 different fables \cite{1834aesop}. We have used these 70 narratives (Table \ref{tab:inputdata}) to investigate the BERT-generated representations of these narratives in the context of semantic content as well as writing style. We have found that writing style is processed in earlier transformer blocks (block: 1) than the semantic content (block: 4-5), which means that the transformer blocks of BERT are specialized to perform certain tasks of language processing. Additionally, we have shown that the clustering of the CLS tokens according to style and content gets better for the last transformer blocks.
The fact that the GDV values increase as a function of the transformer blocks starting from a local minimum value (4-5 for semantics, 1 for style) but decreases again for the last blocks is an interesting finding, which has to be investigated in further studies. Nevertheless, these findings might fit to previous findings by Liu and co-workers (\cite{liu2019linguistic}, explained in detail below).

\subsection*{Relevance of the study in the light of the current literature}
To the best of our knowledge, this study is the first to use a LLM (GPT4) to systematically generate stylistic variations of narratives in order to unravel the mechanisms in the hidden transformer blocks of another LLM (BERT). The relevance of investigating the function of LLMs in order to improve them on the one hand and to use them as a model for the brain on the other hand (see also \cite{lamarre2022attention, kriegeskorte2018cognitive, schilling2023auditory}) is emphasized by the fact that in recent years many high-rank studies occurred trying to unravel the 'black box' of different LLMs (see e.g. \cite{valeriani2024geometry, hewitt2019structural, belinkov2018evaluating, liu2019linguistic, tenney2019bert, rogers2021primer, ma2019universal}). In these studies, the hidden representations of different LLMs and neural machine translation models (NMT) were analyzed regarding semantic \cite{belinkov2018evaluating, liu2019linguistic, tenney2019bert, rogers2021primer} and syntactic processing \cite{hewitt2019structural}. However, in most cases so called "probes" -neural networks specifically trained on decoding information in hidden units of LLMs \cite{conneau2018you, hupkes2018visualisation} - were used to quantify the information content (semantic and syntactic) of the hidden representations \cite{valeriani2024geometry, hewitt2019structural}. Thus, it is sometimes difficult to disentangle the true representations and the content added through the trained probes. In our study, we used exclusively well-defined and established statistical measures such as the generalized discrimination value (GDV, \cite{schilling2021quantifying}) to investigate information processing in the hidden transformer blocks, a fact that limits the comparability between ours and other recent studies. Nevertheless, Liu et al. showed that in contrast to recurrent neural networks in transformer based language models (such as BERT) the transferability of contextual word representations, i.e. the possibility to use the representations of pre-trained LLMs for many different downstream tasks \cite{chiang2022transferability} is best in the middle transformer blocks \cite{liu2019linguistic}. This finding fits to our finding that the GDV for narrative content has a minimum in the middle layers of BERT (Fig. \ref{fig:GDV} d), which means that there is the maximum class separation between the BERT-representations of the different narratives with different semantic content \cite{liu2019linguistic}. The major advantage of our method is that for the calculation of the GDV value the network does not have to be trained on a certain downstream task.

\subsection*{Conclusion}
In summary, we showed an innovative approach to analyzing the processing principles of LLMs using input narratives from another LLM and calculating the GDV for semantic and stylistic differences in certain narratives. Our method is applicable to all pre-trained language models and does not rely on specialized downstream tasks. We hope that the our novel approach of systematically varying narratives using LLMs is useful for further attempts to open the black box of LLMs and to compare there internal mechanisms and principles to those of the human brain, i.e. to follow the route of the cognitive computational neuroscience approach (CCN, \cite{kriegeskorte2018cognitive}).

\section*{Author Contributions}
AS, and PK developed the idea for the study. AS and PK supervised the study. JH and AS programmed the analysis scripts. AS visualized the data. AS, PK, AM, PU and CM wrote the manuscript. All authors discussed the results and approved the final version of the manuscript.

\section*{Data and code availability statement}
The complete data and analysis programs will be made available upon reasonable request.

\section*{Acknowledgments}
This work was funded by the Deutsche Forschungsgemeinschaft (DFG, German Research Foundation): grants KR\,5148/2-1 (project number 436456810), KR\,5148/3-1 (project number 510395418), KR\,5148/5-1 (project number 542747151), and GRK\,2839 (project number 468527017) to PK, and grant SCHI\,1482/3-1 (project number 451810794) to AS. Furthermore, the research leading to these results has received funding from the European Research Council (ERC) under the European Union’s Horizon 2020 research and innovation programme (ERC Grant No. 810316 to AM).

\bibliographystyle{IEEEtran}
\bibliography{references}
\end{document}